\documentclass[10pt,twocolumn,letterpaper]{article}

\usepackage{cvpr}              % To produce the CAMERA-READY version
\usepackage{algorithm}
\usepackage{algorithmic}

\usepackage{amsmath,amsfonts}
\usepackage{url}
\usepackage{booktabs} 
\usepackage{multicol}
\usepackage{multirow}
\usepackage{pifont}       % \ding{xx}
\usepackage{bbding}       % \Checkmark,\XSolid,... (需要和pifont宏包共同使用)
\usepackage{fontawesome}  % \faCheck,\faTimes

\definecolor{cvprblue}{rgb}{0.21,0.49,0.74}
\usepackage[pagebackref,breaklinks,colorlinks,allcolors=cvprblue]{hyperref}

\def\paperID{8487} % *** Enter the Paper ID here
\def\confName{CVPR}
\def\confYear{2026}

\title{DMAligner: Enhancing Image Alignment via Diffusion Model \\ Based View Synthesis}

\author{
Xinglong Luo\textsuperscript{\rm 1, \rm 2},
Ao Luo\textsuperscript{\rm 3},
Zhengning Wang\textsuperscript{\rm 1}\footnotemark[2], \\
Yueqi Yang\textsuperscript{\rm 4},
Chaoyu Feng\textsuperscript{\rm 4},
Lei Lei\textsuperscript{\rm 4},
Bing Zeng\textsuperscript{\rm 1},
and Shuaicheng Liu\textsuperscript{\rm 1}\footnotemark[2]\\
\textsuperscript{\rm 1}University of Electronic Science and Technology of China \\
\textsuperscript{\rm 2}Qianyuan Laboratory, Hangzhou, China
\textsuperscript{\rm 3}Southwest Jiaotong University 
\textsuperscript{\rm 4}Independent Researcher\\
\tt\small \{xinglong.luo@std.,zhengning.wang@,eezeng@,liushuaicheng@\}uestc.edu.cn
\tt\small aoluo@swjtu.edu.cn
}

\begin{document}
\maketitle
\renewcommand{\thefootnote}{\fnsymbol{footnote}}
% \footnotetext[1]{Equal contribution.}\footnotetext[2]{Corresponding Author.}
\footnotetext{\footnotemark[2] Corresponding Author.}

\begin{abstract}
Image alignment is a fundamental task in computer vision with broad applications. Existing methods predominantly employ optical flow-based image warping. However, this technique is susceptible to common challenges such as occlusions and illumination variations, leading to degraded alignment visual quality and compromised accuracy in downstream tasks. 
In this paper, we present DMAligner, a diffusion-based framework for image alignment through alignment-oriented view synthesis. DMAligner is crafted to tackle the challenges in image alignment from a new perspective, employing a generation-based solution that showcases strong capabilities and avoids the problems associated with flow-based image warping. 
Specifically, we propose a Dynamics-aware Diffusion Training approach for learning conditional image generation, synthesizing a novel view for image alignment. 
This incorporates a Dynamics-aware Mask Producing (DMP) module to adaptively distinguish dynamic foreground regions from static backgrounds, enabling the diffusion model to more effectively handle challenges that classical methods struggle to solve.
Furthermore, we develop the Dynamic Scene Image Alignment (DSIA) dataset using Blender, which includes 1,033 indoor and outdoor scenes with over 30K image pairs tailored for image alignment.
Extensive experimental results demonstrate the superiority of the proposed approach on DSIA benchmarks, as well as on a series of widely-used video datasets for qualitative comparisons. Our code is available at \url{https://github.com/boomluo02/DMAligner}.
\end{abstract}

\section{Introduction}

The alignment of consecutive frames is a critical task in computer vision, supporting a wide range of applications such as High-Dynamic-Range (HDR) imaging~\cite{xu2024hdrflow}, image deblurring~\cite{Youk_2024_CVPR}, burst image restoration~\cite{dudhane2022burst}, under-display camera imaging~\cite{ding2025Neurocomputing}, video super-resolution~\cite{chan2021understanding}, video denoising~\cite{yue2020supervised} and video stabilization~\cite{liu2025minimum}. Precise alignment ensures the accurate merging of information from consecutive frames, leading to enhanced image quality and better performance in subsequent tasks~\cite{pan2024burst,wang2024reconstruction}. Despite its significance, the task of image alignment remains challenging due to various factors, including scene dynamics, large camera motion, and varying lighting conditions.

\begin{figure}[t]
    \centering
    \includegraphics[width=.98\linewidth]{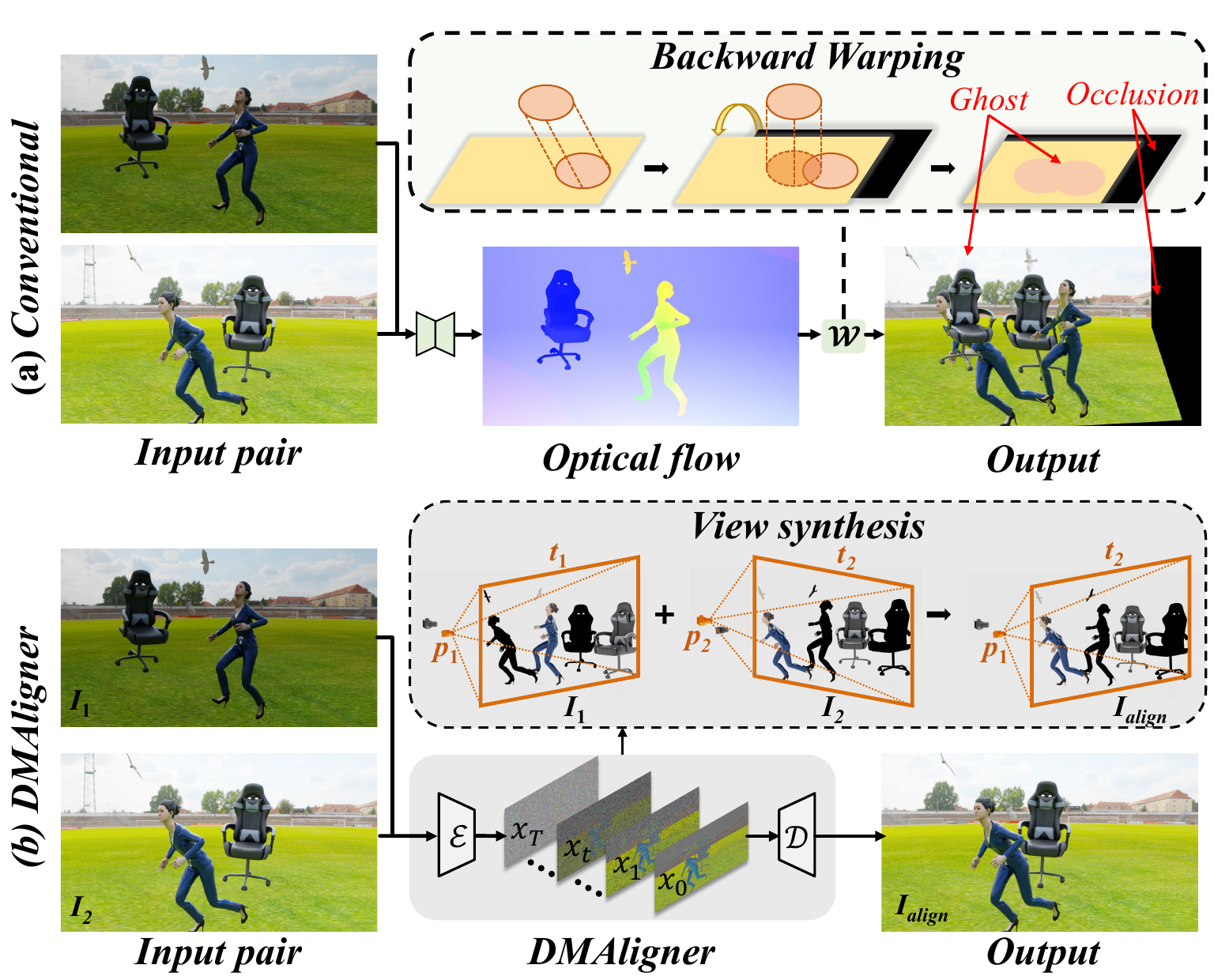}
    \caption{(a) Conventional image alignment based on optical flow and image warping, resulting in ghosting artifacts and occlusion. (b) Our DMAligner directly generates the complete alignment image via diffusion-based view synthesis.}
    \label{fig:teaser}
    \vspace{-1em}
\end{figure}

Recent advancements in deep learning have led to substantial progress in neural network-based alignment methods. 
Generally, image alignment can be divided into three levels: global-level, mid-level, and pixel-level alignment. Global-level alignment typically involves estimating a homography matrix, which can be accomplished using unsupervised deep learning frameworks~\cite{liu2022content}. This technique effectively aligns images under conditions of planar scenes or when the camera undergoes basic rotational or translational motion.
Moreover, through global-to-local homography flow refinement, it can be naturally extended to local mesh-grid homography estimation~\cite{liu2022unsupervised}, allowing it to overcome the limitations of a single homography~\cite{liu2022content}.
However, global- and mid-level methods are restricted in their ability to address more complex scenarios, like parallax, non-rigid deformations, or significant foreground motion, since they assume single or multiple global transformations across the entire image.

Another prominent technique involves the pixel-level approaches with optical flow and image warping~\cite{bhat2021deep1}. Despite progress in image alignment, current optical flow-based methods face challenges that limit their effectiveness in practical applications. A major issue is the sensitivity of optical flow methods to brightness changes. Variations in lighting conditions, such as shadows, reflections, or changes in ambient light, can significantly distort the brightness constancy assumption, leading to incorrect motion estimation. Besides, the warping process inherent in optical flow-based alignment can introduce various artifacts. Inaccuracies in the motion field can lead to distortions such as stretching, compressing, or tearing of image regions. Additionally, the many-to-one mapping inherent in image warping on occlusion regions inevitably results in ghosting artifacts, as depicted in Fig.~\ref{fig:teaser} (a), even with precise optical flow estimation. 
Such artifacts not only degrade the visual quality of the aligned frames but also affect the accuracy of subsequent processing tasks.

\begin{figure}[t]
    \centering
    \includegraphics[width=\linewidth]{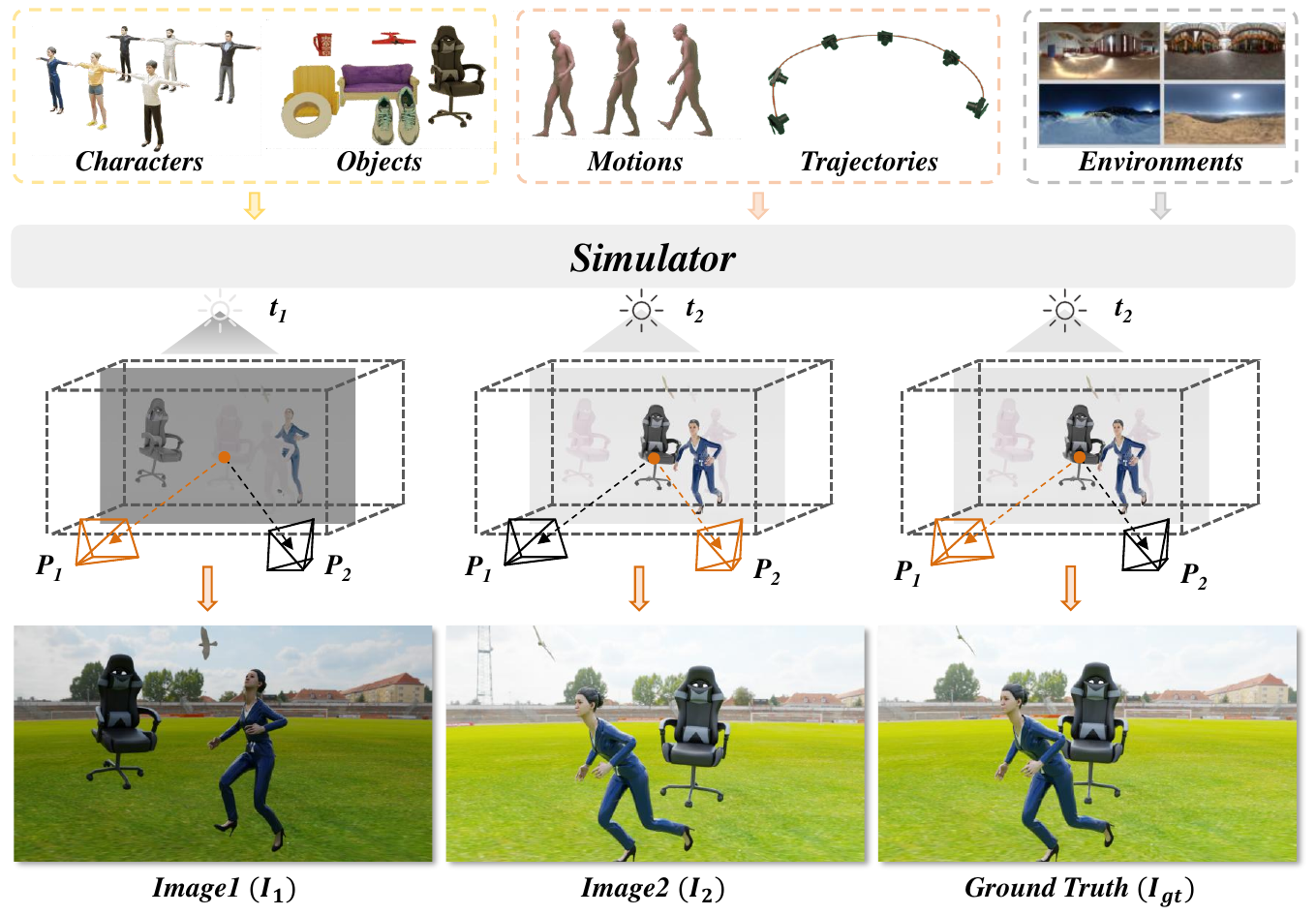}
    \caption{Overview of DSIA dataset generation. The ground truth image is rendered by setting the time to $t_2$ and the camera pose to $P_1$, where the background statics mirror those in $I_1$, and the dynamics are akin to $I_2$, with slight visualization variances caused by pose changes. As a result, $I_{gt}$ can be regarded as the {\em alignment-oriented view synthesis} with reference from $I_2$ to $I_1$ ({\em i.e.,} ${I_2}'$ for $I_2$ to $I_1$ alignment).}
    \label{fig:datapipeline}
    \vspace{-.5em}
\end{figure}

In this paper, we introduce DMAligner, a diffusion-based framework for image alignment through alignment-oriented view synthesis. 
As depicted in Fig.~\ref{fig:datapipeline}, the desired image alignment result represents a scenario where the camera pose stays fixed at $P_1$ while time advances from $t_1$ to $t_2$.
DMAligner is designed to handle the challenges in image alignment from a new perspective with a generation-based solution (see Fig.~\ref{fig:teaser} (b)), exhibiting strong capabilities and avoiding issues associated with homography and flow-based image warping. 
Specifically, we first create Dynamic Scene Image Alignment (DSIA) dataset, designed to simulate typical image alignment challenges including camera shifts, moving objects, and fluctuating illumination. 
% The dataset comprises 1,033 scenes from both indoor and outdoor environments, with more than 30,000 image pairs, all at a resolution of 960$\times$540.
Moreover, we propose the Dynamics-aware Diffusion Training approach on the Latent Diffusion Model (LDM)~\cite{rombach2022ldm} for learning conditional image generation.
% , where a new image is synthesized as the result of image alignment. 
The key idea is to enhance the network's capacity to recognize dynamic information during diffusion model training. This is realized through the Dynamics-aware Mask Producing (DMP) module, which effectively discriminates dynamic foreground regions from static backgrounds within the hidden features across frames.
Additionally, we validate the effectiveness of DMAligner through extensive testing on synthetic datasets, including the proposed DSIA and Sintel~\cite{butler2012sintel}, as well as on the widely-used real-world dataset DAVIS~\cite{caelles2019DAVIS}. Both the quantitative and qualitative results demonstrate significant enhancements over contemporary methods. The contributions are summarized as follows:

\begin{itemize}

\item We develop {\bf DSIA} dataset, the first large-scale dataset specific for image alignment, which simulates the typical challenges including camera motion, moving objects, and changes in illumination.

\item We present {\bf DMAligner}, a novel diffusion-based framework for image alignment via alignment-oriented view synthesis. The central idea is to enhance the network's capability to capture dynamic information during diffusion model training, with the assistance of our {\bf DMP} module. 

\item The proposed method achieves {\bf state-of-the-art performance} on DSIA benchmarks and also demonstrates the {\bf superior generalization capability} on both synthetic and real-world datasets. 

\end{itemize}

\section{Related Work}
\paragraph{Image Alignment.} 
Traditional approaches primarily rely on feature-based methods, where keypoints are detected and matched across images, followed by geometric transformation estimation to align the images~\cite{liu2010sift}.
In the deep learning era, deep convolutional neural networks (DCNNs) have revolutionized image alignment in feature space~\cite{Liu_2021_HDRDCNN}. Furthermore, the Transformer has also been utilized for feature-domain alignment, extending their application beyond standard RGB images~\cite{chen2023improving} to directly process RAW image data~\cite{ren2025ISPFormer,ren2025learning}.
% In the deep learning era, DCNNs~\cite{zhu2019deformable} have revolutionized image alignment in feature space, which is widely applied in video super-resolution~\cite{tian2020tdan}, video denoising~\cite{yue2020supervised}, and burst image restoration~\cite{dudhane2022burst}.
In image space, Global- and middle-level alignments typically involve estimating a homography matrix~\cite{liu2022content} and local mesh-grid homography~\cite{liu2022unsupervised}, respectively. But they cannot handle complex scenarios, involving parallax, non-rigid deformations, or significant foreground motion.
A notable advancement in image alignment is the adoption of optical flow-based image warping techniques. Optical flow calculates pixel-wise motion between consecutive images, enabling the warping of one image to align with another. With the advances of recent optical flow approaches~\cite{luo2023gaflow,deng2023emdflow,Luo_2023_HREM}, this method has found widespread use across a variety of video-based applications~\cite{liu2022hdr,zhang2021deep}.
Despite their success, warping in occlusion regions inevitably leads to ghosting artifacts, even with accurate flow, as shown in Fig.~\ref{fig:teaser}. In contrast, our DMAligner addresses the challenges in image alignment from a new perspective, performing {\em alignment-oriented view synthesis} using a diffusion-based framework.

\begin{figure*}[t]
    \centering
    \includegraphics[width=.96\linewidth]{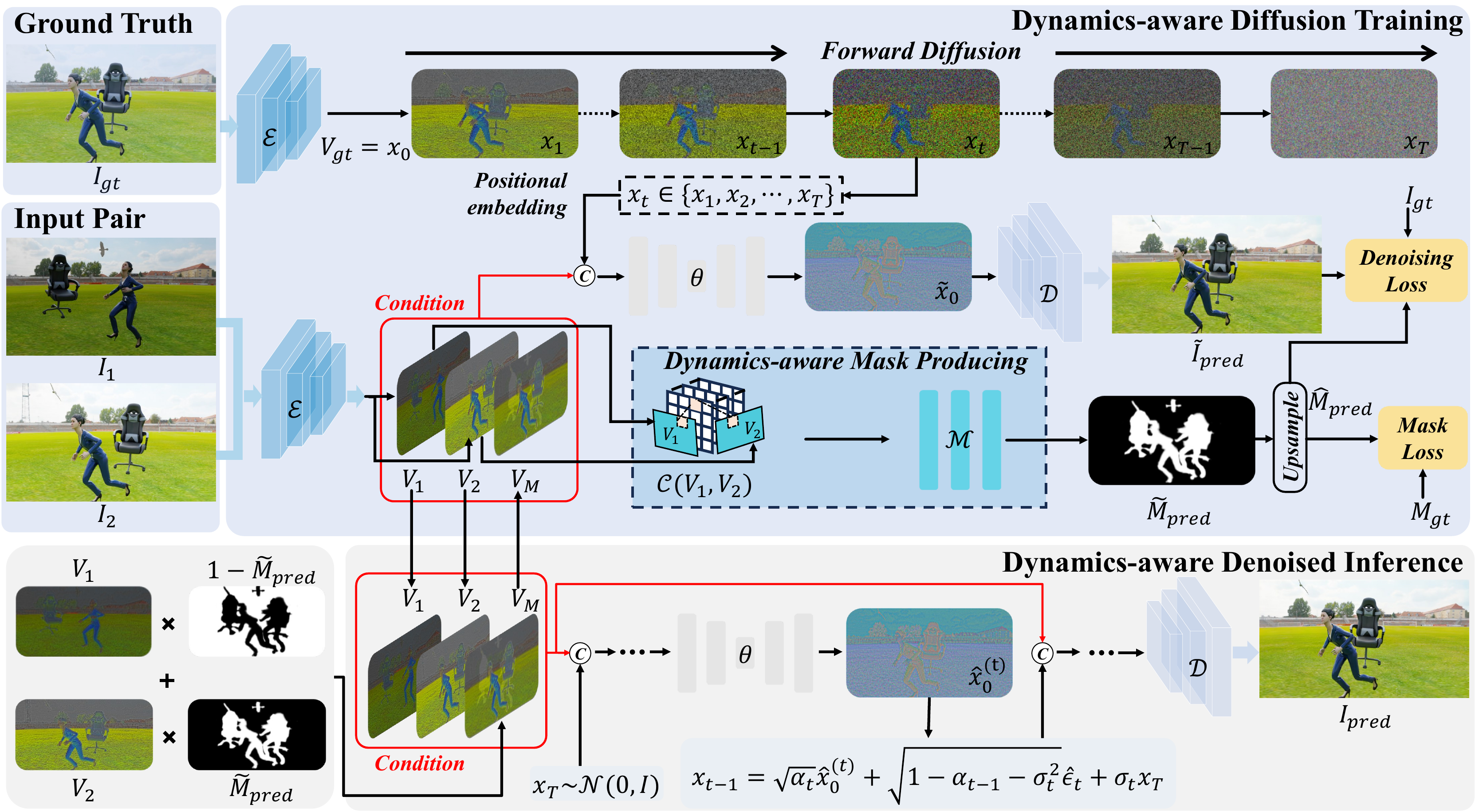}
    \vspace{-.5em}
    \caption{Overview of our DMAligner. Instead of using the discriminative learning paradigm for optical flow estimation and image warping, our framework employs a generative approach to achieve image alignment with a diffusion model. Dynamics-aware Mask Producing (DMP) module is crucial for providing dynamic information, essential for performing the Dynamics-aware Diffusion Training process in this task.}
    \label{fig:network}
    \vspace{-.5em}
\end{figure*}

\paragraph{Diffusion Model.} 
Diffusion models~\cite{ho2020ddpm,song2021ddim,rombach2022ldm} own prominent generative performance for synthesis, and various approaches have been developed to improve the performance or efficiency. Different applications have shown good performances when empowered with diffusion models, including but not limited to, optical flow estimation~\cite{Luo2024FlowDiffuser},  homography estimation~\cite{li2024dmhomo}, shadow removal~\cite{luo2025diff}, image rectangling~\cite{zhou2024recdiffusion}, rolling shutter removal~\cite{yang2025single}, HDR image reconstruction~\cite{liu2025solvingHDR}, low-light image enhancement~\cite{jiang2023low, jiang2024lightendiffusion}, raw image processing~\cite{liu2026RAWFlow,ren2025ispdiffuser}, camouflaged object detection~\cite{zhao2024focusdiffuser} and robot manipulation~\cite{chi2023diffusionpolicy,zhang2025flowpolicy,xu2026hero,xu2026AGPred}. In this work, we propose an end-to-end diffusion-based framework that {\em specifically handles cross-frame dynamic information} for image alignment.

\section{Method}
\subsection{DSIA Dataset Generation}
% We have developed a novel paradigm for image alignment by leveraging diffusion models to produce images of objects captured from desired viewpoints in accurate poses. Additionally, we have constructed a data generation platform based on the rendering engine, capable of synthesizing data in batches. Using this platform, we have created the first synthetic viewpoint image alignment dataset, which we refer to as the SVIA dataset. 
To the best of our knowledge, {\em there is no large-scale dataset that addresses the specific challenges of image alignment, such as camera motion, moving objects, and varying illumination, while also providing corresponding ground truth images}. To fill this gap, we create the Dynamic Scene Image Alignment (DSIA) dataset, which simulates these challenges and includes 1,033 indoor and outdoor scenes with over 30K image pairs at a resolution of 960$\times$540. 
The main idea is to {\em simulate the burst image capturing processes with the moving camera across a variety of dynamic scenes}.
Fig.~\ref{fig:datapipeline} illustrates the overview of DSIA dataset generation.

Specifically, inspired by PointOdyssey~\cite{zheng2023pointodyssey} and Kubric~\cite{greff2022kubric}, we assemble 25 characters and 100 objects with diverse shapes and materials from several open-source sources, including BlenderKit, Mixamo, GSO, and PartNet. These characters are animated using real-world motion capture data, while the locations and rotations of object poses are randomly adjusted to simulate real-world motion. Camera motion is controlled along predefined trajectories extracted from~\cite{Zhengqi_characters}. By collecting and animating these assets (characters, objects, and cameras), we generate 3D virtual scenes that closely mimic real-world dynamics. These scenes are then rendered with the Blender engine, producing image sequences $\{I_1, I_2, \cdots, I_N\}$, in which characters, objects, and the camera all move simultaneously. To maximize the diversity of the generated images, we randomly introduce HDR environments with various materials, textures, and lighting conditions.

We systematically rearrange the sequencing of the camera trajectories relative to character and object motions, capturing dynamic scenes from various positions along the trajectories, and rendering to generate virtual viewpoint images tailored for image alignment. 
% As shown in Fig.~\ref{fig:datapipeline}, 
Considering two consecutive images, $I_1$ and $I_2$ involve both global camera motion $\mathcal{H}$ and local foreground movement $\mathcal{F}$. Their transformations can be described using the traditional warp operation $\mathcal{W}$:
\begin{equation}
I_1=\mathcal{W}(I_2)+\mathcal{T}=\mathcal{H}(\mathcal{F}(I_2))+\mathcal{T},
\label{I2toI1}
\end{equation}
where $\mathcal{T}$ represents the changes in image texture. In our approach, we capture the dynamic scene at the time of $I_2$ from the camera viewpoint of $I_1$, generating the aligned image $I_{gt}$ as Ground Truth. Compared to $I_1$, $I_{gt}$ involves local foreground movement $\mathcal{F}$, while compared to $I_2$, $I_{gt}$ involves global camera motion $\mathcal{H}$:
\begin{equation}
I_{gt}=\mathcal{H}(I_2)=\mathcal{F}(I_1)+\mathcal{T}.
\label{IgtI2I1}
\end{equation}
The dataset is divided into four subsets based on varying camera and foreground movements.

\subsection{Diffusion Model for Image Alignment}

\subsubsection{Overview.} 
The training and inference pipelines of our DMAligner are depicted in Fig.~\ref{fig:network}. The entire process is formulated in the latent space using LDM~\cite{rombach2022ldm}. The core insight is to equip the network with the ability to capture dynamic information during diffusion model training. This is accomplished by the Dynamics-aware Mask Producing (DMP) module, which effectively differentiates dynamic foreground regions from static backgrounds within the cross-frame hidden features. We elaborate on the model training and inference details in the following subsections.

\subsubsection{Dynamics-aware Diffusion Training.}
Given a set of image data $I_1$, $I_2$, and $I_{gt}$, we follow LDM~\cite{rombach2022ldm} to transform these high-resolution images into the latent representational space using a perceptual compression model based on the encoder-decoder architecture, and then reconstruct them back into RGB images. Specifically, for $I_{gt}$, the encoder $\mathcal{E}$ encodes it into the latent representation $V_{gt}$, and the decoder $\mathcal{D}$ reconstructs it into the RGB image: 
\begin{equation}
\tilde{I}_{gt} = \mathcal{D}(V_{gt}) = \mathcal{D}(\mathcal{E}(I_{gt})). 
\label{vae}
\end{equation}
The same process is applied to $I_1$ and $I_2$, resulting in latent representations $V_1$ and $V_2$. We use $V_1$ and $V_2$ as the condition in the latent representational space and train the diffusion model $\theta$ to learn sampling from standard Gaussian noise, generating the latent that matches the distribution of $V_{gt}$. Our training scheme consists of four components: Forward Diffusion, Conditional Learning, Dynamics-aware Mask Producing, and Loss Function.

\textbf{Forward Diffusion:} As illustrated in Fig.~\ref{fig:network}, we apply different timesteps $t$ to add varying levels of noise to $V_{gt}$ (denoted as $x_0$), generating a series of noise representations $x_t \in \{x_1, x_2, \cdots, x_T\}$. $x_t$ indicates any noise representation at a specific timestep $t$, and as $t$ increases, the level of added noise grows, leading to an increase in the entropy of the distribution of $x_t$. Ultimately, the distribution becomes standard Gaussian noise, $x_T \sim \mathcal{N}(0, 1)$, defined as:
\begin{equation}
x_t \in \{x_1,x_2,\cdots,x_T\},\ x_T\sim \mathcal{N}(0,1),\ x_0=V_{gt}.
\label{forwardxt}
\end{equation}
The forward diffusion process typically follows the approach of \cite{ho2020ddpm}, where Gaussian noise is incrementally added to the data sample based on a Markov chain. This allows for the diffusion from $x_0$ to any timestep $x_t$ in one step using the following equation:
\begin{equation}
q(x_t \mid x_0)=\mathcal{N}(x_t \mid \sqrt{\bar{\alpha}_t} x_0,(1-\bar{\alpha}_t)I),
\label{forwarddiffusion}
\end{equation}
where $\bar{\alpha}_t$ represents the cumulative product of a set of pre-defined noise variance schedules $\{ \beta_1, \beta_2,\cdots, \beta_t \}$, which ensures that the noise incrementally increases with $t$, defined as $\bar{\alpha}_t : =   {\textstyle \prod_{i=1}^{t}} \alpha _{i}={\textstyle \prod_{i=1}^{t}} (1-\beta_i)$.

\textbf{Dynamics-aware Mask Producing:}
To explicitly guide the denoising network to focus on moving foreground objects and enhance local detail reconstruction, we utilize a motion mask. First, we extract motion information by computing a normalized dot-product cost volume $\mathcal{C}$ between latents $V_1$ and $V_2$. For a spatial coordinate $\mathbf{u}$ and a displacement $\mathbf{d}$, the correlation is defined as:
\begin{equation}
\mathcal{C}\{\mathbf{u}, \mathbf{d}\} = \frac{1}{N}\sum_{k=1}^{N} V_1^{(k)}(\mathbf{u}) \cdot V_2^{(k)}(\mathbf{u} + \mathbf{d}),
\end{equation}
where $N$ is the channel dimension. Assuming the input images $I_1$ and $I_2$ have a spatial resolution of $\mathbb{R}^{3 \times H \times W}$ and the encoder $\mathcal{E}$ has a downsampling factor of $s$, the resulting latents $V_1$ and $V_2$ both belong to $\mathbb{R}^{N \times \frac{H}{s} \times \frac{W}{s}}$. Consequently, the correlation volume $\mathcal{C} \in \mathbb{R}^{D^2 \times \frac{H}{s} \times \frac{W}{s}}$ (where $D$ represents the maximum displacement) is fed into a mask predictor $\mathcal{M}$ to generate the motion mask $\tilde{M}_{pred} = \mathcal{M}(\mathcal{C})$. To ensure smooth visual transitions at object boundaries, we expand this mask using a dilation operation $d_r$ (which is fully differentiable and implemented via the PyTorch ``torch.nn.functional.max\_pool2d'' operation with a stride of 1). The dilated mask is subsequently used to fuse the background of $V_1$ with the foreground of $V_2$:
\begin{equation}
V_M = V_2 \odot d_r(\tilde{M}_{pred}) + V_1 \odot \{1 - d_r(\tilde{M}_{pred})\}.
\end{equation}
This mixed latent $V_M$ serves as a critical conditioning signal to guide the diffusion network in generating the final aligned output.

\textbf{Conditional Learning:} 
As shown in Fig.~\ref{fig:network}, we concatenate $V_1$, $V_2$ and $V_M$ along the channel dimension to form the condition $x_{cond}$, then randomly select $x_t$ from the noise representations $\{x_1, x_2, \cdots, x_T\}$, allowing the diffusion model $\theta$ to learn the denoising process. This enables $\theta$ to handle any $x_t$ at an arbitrary timestep $t$, producing a denoised output $\hat{x}_0$. Similar to approaches in~\cite{Luo2024FlowDiffuser}, we use the ``predict $x_0$" denoising setting in our model. The condition $x_{cond}$ is concatenated with $x_t$ and the encoded timestep $t$ through positional embedding (PE), and the resulting tensor is sent to the diffusion model to obtain the denoised result $\tilde{x}_0$:
\begin{equation}
\tilde{x}_0 = \theta([x_t,x_{cond},{\rm PE}(t)]),
\label{forwardpredx0}
\end{equation}
then $\tilde{x}_0$ is reconstructed to become $\tilde{I}_{pred}$ by decoder $\mathcal{D}$, which can be expressed as $\tilde{I}_{pred}=\mathcal{D}(\tilde{x}_0)$. Experiments demonstrate that ``predict $x_0$" is more effective for image alignment compared to others.
\begin{figure*}[t]
    \centering
    \includegraphics[width=1\linewidth]{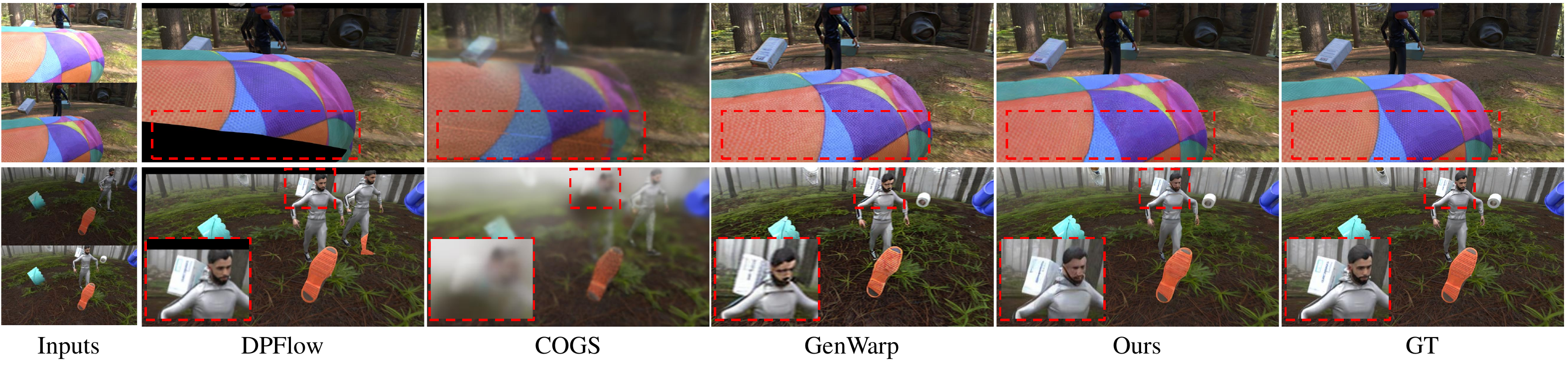}
    \vspace{-2em}
    \caption{Qualitative comparisons between DPFlow~\cite{Morimitsu2025DPFlow}, COGS~\cite{jiang2024cogs}, GenWarp~\cite{seo2024genwarp} and our DMAligner on our DSIA dataset.}
    \label{fig:comparison}
    \vspace{-.6em}
\end{figure*}

\begin{table}[t]
\centering
\small
\caption{Comparison of network inputs among DPFlow~\cite{Morimitsu2025DPFlow}, COGS~\cite{jiang2024cogs}, AccidentalGS~\cite{mao2025accidentalgs}, GenWarp~\cite{seo2024genwarp} and our DMAligner. Other optical flow networks (PWCNet~\cite{sun2018pwcnet}, RAFT~\cite{teed2020raft}, FlowFormer++~\cite{shi2023flowformer++}, FlowDiffuser~\cite{Luo2024FlowDiffuser}) have the same input structure as DPFlow.}
\vspace{-1.em}
\begin{tabular}{l|ccccccc}
\toprule
\textbf{Method} & 
\textbf{Input Structure} \\
\hline
{DPFlow~\cite{Morimitsu2025DPFlow}} 
& 2 images ($I_1, I_2$) \\
\hline
\multirow{2}{*}{{COGS~\cite{jiang2024cogs}}}
& 2 images ($I_1, I_2$) + \\
& Depth ($D_1, D_2$) + Mask ($M_1, M_2$) \\
\hline
\multirow{2}{*}{{AccidentalGS~\cite{mao2025accidentalgs}}}
& 2 images ($I_1, I_2$) + \\
& Depth ($D_1, D_2$) + Camera Intrinsics ($K$) \\
\hline
\multirow{2}{*}{{GenWarp~\cite{seo2024genwarp}}}
& Single image ($I_2$) + Camera Intrinsics ($K$) + \\
& Depth ($D_2$) + Camera Motion ($P_{I_2\to I_1}$)\\
\hline
{DMAligner} 
& 2 images ($I_1, I_2$)\\
\bottomrule
\end{tabular}
\vspace{-1.5em}
\label{tab:input_comparison}
\end{table}

\begin{table*}
  \centering
    \caption{Quantitative comparison on our DSIA dataset across four subsets (LcLf, LcSf, ScLf, and ScSf) using PSNR and SSIM metrics, \textit{higher scores indicate better performance}. The average (``Avg") across all subsets summarizes overall results. $^{\dagger}$ indicates that occlusion and ghost regions are excluded during metric computation, as determined by forward-backward consistency check~\cite{xu2022gmflow}.}
    \vspace{-.6em}
    \begin{tabular}{l|rr|rr|rr|rr|rr}
    \toprule
    \multirow{2}{*}{\textbf{Methods}} & \multicolumn{2}{c|}{\textbf{LcLf}} & \multicolumn{2}{c|}{\textbf{LcSf}} & \multicolumn{2}{c|}{\textbf{ScLf}} & \multicolumn{2}{c|}{\textbf{ScSf}} & \multicolumn{2}{c}{\textbf{Avg.}} \\
         & PSNR$\uparrow$ & SSIM$\uparrow$ & PSNR$\uparrow$ & SSIM$\uparrow$ & PSNR$\uparrow$ & SSIM$\uparrow$ & PSNR$\uparrow$ & SSIM$\uparrow$ & PSNR$\uparrow$ & SSIM$\uparrow$ \\
    \hline
    \multicolumn{11}{c}{\textit{Methods based on Flow Warping}} \\  % 合并两列并居中
    \hline
    PWCNet~\cite{sun2018pwcnet} & 16.53  & 0.74  & 16.40  & 0.73  & 20.53  & 0.74  & 21.01  & 0.76  & 18.62  & 0.74  \\
    RAFT~\cite{teed2020raft} & 18.22  & 0.74  & 18.34  & 0.74  & 22.46  & 0.77  & 22.95  & 0.76  & 20.49  & 0.75  \\
    FlowFormer++~\cite{shi2023flowformer++} & 21.02  & 0.75  & 20.97  & 0.76  & 22.84  & 0.77  & 23.94  & 0.79  & 22.19  & 0.77  \\
    FlowDiffuser~\cite{Luo2024FlowDiffuser} & 20.18  & 0.75  & 20.33  & 0.75  & 23.39  & 0.78  & 23.91  & 0.79  & 21.95  & 0.77  \\
    DPFlow~\cite{Morimitsu2025DPFlow} & 20.78  & 0.76  & 20.50  & 0.74  & 23.94  & 0.78 & 24.61  & 0.80 & 22.46  & 0.77 \\
    \hline
    PWCNet$^{\dagger}$~\cite{sun2018pwcnet} & 22.03  & 0.76  & 21.15  & 0.76  & 24.63  & 0.79  & 24.74  & 0.77  & 23.14  & 0.77  \\
    RAFT$^{\dagger}$~\cite{teed2020raft}  & 22.99  & 0.78  & 23.15  & 0.77  & 25.07  & 0.80  & 25.09  & 0.79  & 24.08  & 0.78  \\
    FlowFormer++$^{\dagger}$~\cite{shi2023flowformer++} & 24.19  & 0.79  & 24.24  & 0.78  & 25.61  & 0.81  & 25.74  & 0.81  & 24.95  & 0.80  \\
    FlowDiffuser$^{\dagger}$~\cite{Luo2024FlowDiffuser} & 23.82  & 0.78  & 23.81  & 0.77  & 25.78  & 0.81  & 25.89  & 0.82  & 24.82  & 0.80  \\
    DPFlow$^{\dagger}$~\cite{Morimitsu2025DPFlow} & \underline{24.30}  & \underline{0.79}  & \underline{24.84}  & \underline{0.79}  & \underline{25.99}  & \textbf{0.82} & \underline{26.04}  & \underline{0.82} & \underline{25.29}  & \underline{0.81} \\
    \hline
    \multicolumn{11}{c}{\textit{Methods based on View Synthesis}} \\  % 合并两列并居中
    \hline
    COGS~\cite{jiang2024cogs} & 16.35  & 0.66  & 16.84  & 0.69  & 17.01  & 0.69  & 18.21  & 0.71  & 17.10  & 0.69 \\
    AccidentalGS~\cite{mao2025accidentalgs} & 17.47 & 0.68 & 19.12 & 0.70 & 20.38 & 0.71 & 21.33 & 0.74 & 19.58 & 0.71 \\
    SD-Inpaint~\cite{yu2023SDinpainting} & 22.14  & 0.75  & 22.92  & 0.76  & 25.42  & 0.79  & 25.77  & 0.81  & 24.06  & 0.78 \\
    GenWarp~\cite{seo2024genwarp} & 22.62  & 0.76  & 23.15  & 0.76  & 25.32  & 0.78  & 25.84  & 0.81  & 24.23  & 0.78  \\
    \textbf{DMAligner} (Ours) & \textbf{24.91} & \textbf{0.80} & \textbf{26.90} & \textbf{0.81} & \textbf{27.25} & \underline{0.81} & \textbf{27.64} & \textbf{0.83} & \textbf{26.67} & \textbf{0.81}\\
    \bottomrule
    \end{tabular}
  \label{tab:compair}
  \vspace{-1.5em}
\end{table*}

\begin{figure*}[t]
    \centering
    \includegraphics[width=.98\linewidth]{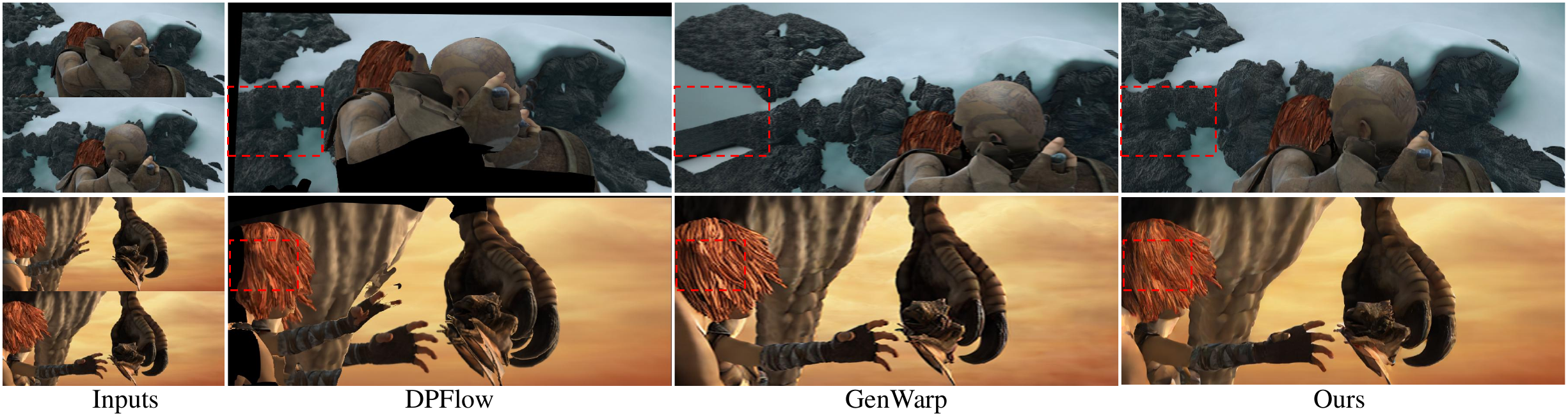}
    \vspace{-1em}
    \caption{Qualitative comparisons between DPFlow~\cite{Morimitsu2025DPFlow}, GenWarp~\cite{seo2024genwarp} and our DMAligner on Sintel dataset.}
    \label{fig:comparison_sintel}
    \vspace{-1.em}
\end{figure*}
\begin{figure*}[t]
    \centering
    \includegraphics[width=.98\linewidth]{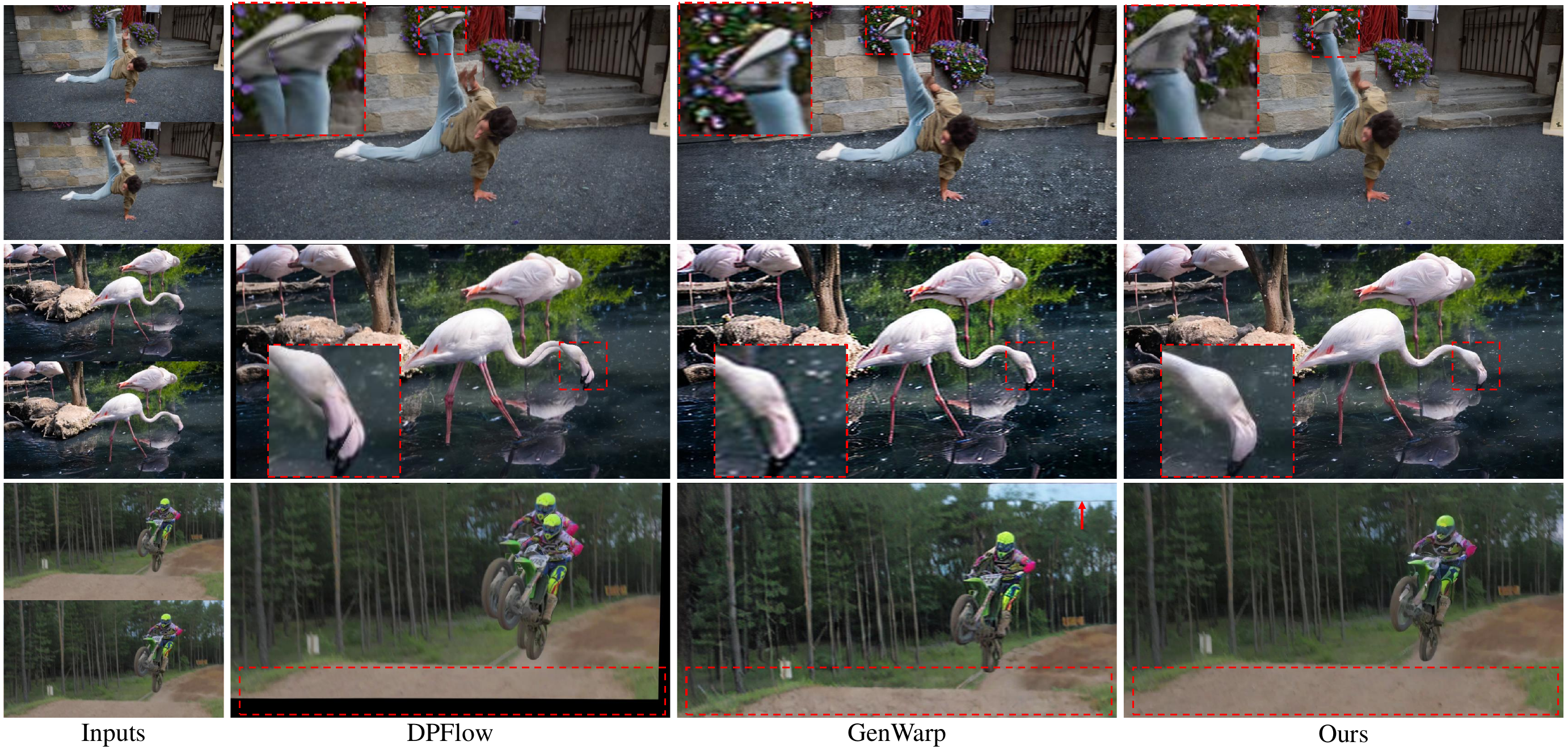}
    \vspace{-1em}
    \caption{Qualitative comparisons between DPFlow~\cite{Morimitsu2025DPFlow}, GenWarp~\cite{seo2024genwarp} and our DMAligner on the real-world DAVIS dataset.}
    \label{fig:comparison_davis}
    \vspace{-1.5em}
\end{figure*}

\begin{figure*}[t]
    \centering
    \includegraphics[width=\linewidth]{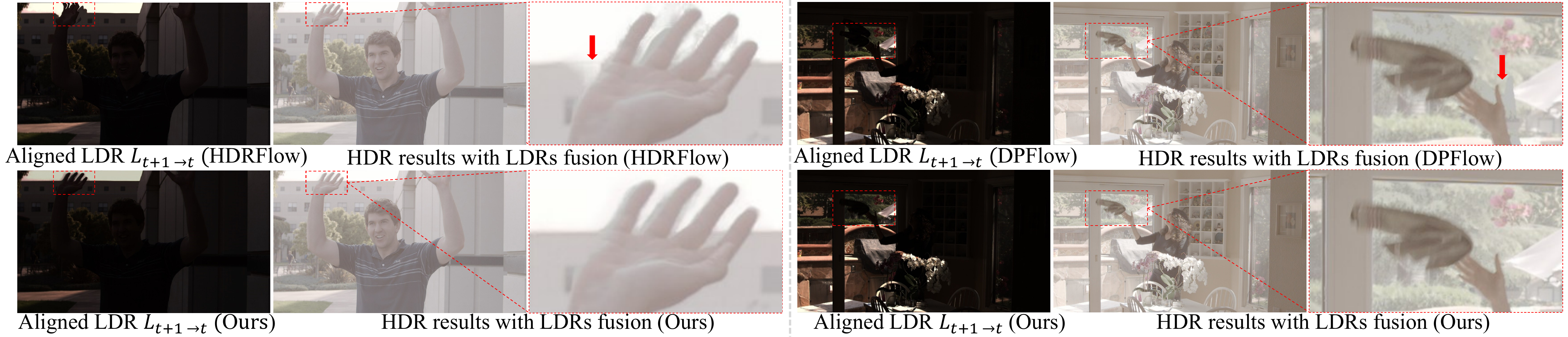}
    \vspace{-0.8cm}
    \caption{Comparison of HDR results using flow-based method HDRFlow~\cite{xu2024hdrflow}, DPFlow~\cite{Morimitsu2025DPFlow} and our DMAligner. Our DMAligner significantly reduces ghosting and background distortions, especially around occluded regions and moving objects.}
\vspace{-0.3cm}
\label{fig:hdr_compairison}
\end{figure*}

\begin{table*}[t]
  \caption{Quantitative comparison on Sintel and DAVIS dataset. Due to the lack of ground truth for precise evaluation, we compute LPIPS and DreamSim~\cite{fu2023dreamsim} between $I_{pred}$ and $I_1$. \textit{Lower scores indicate better performance}.}
\vspace{-1.0em}
  \centering
    \begin{tabular}{l|cc|cc|cc|cc}
    \toprule
    \multirow{2}{*}{\textbf{Methods}} & \multicolumn{2}{c|}{\textbf{Sintel (Clean)}} & \multicolumn{2}{c|}{\textbf{Sintel (Final)}}  & \multicolumn{2}{c|}{\textbf{DAVIS}} &\multicolumn{2}{c}{\textbf{Avg.}} \\
    & LPIPS$\downarrow$      & DreamSim$\downarrow$      & LPIPS$\downarrow$      & DreamSim$\downarrow$        & LPIPS$\downarrow$      & DreamSim$\downarrow$  & LPIPS$\downarrow$      & DreamSim$\downarrow$\\
    \hline
    \multicolumn{9}{c}{\textit{Methods based on Flow Warping}} \\  % 合并两列并居中
    \hline
    PWCNet~\cite{sun2018pwcnet} & 0.251  & 0.146  & \underline{0.177}  & 0.109  & 0.217  & 0.102  & 0.215  & 0.119  \\
    RAFT~\cite{teed2020raft}  & \textbf{0.245} & {0.141}  & \textbf{0.171} & \underline{0.102} & 0.219  & 0.104  & \underline{0.212}  & 0.116  \\
    FlowFormer++~\cite{shi2023flowformer++} & 0.264  & 0.150  & 0.184  & 0.113  & 0.224  & 0.106  & 0.224  & 0.123  \\
    FlowDiffuser~\cite{Luo2024FlowDiffuser} & 0.258  & 0.141  & 0.179  & {0.103}  & {0.211}  & {0.090}  & 0.216  & {0.111}  \\
    DPFlow~\cite{Morimitsu2025DPFlow} & 0.255  & \underline{0.138}  & 0.178  & \textbf{0.101}  & \underline{0.209}  & \underline{0.087}  & 0.214  & \underline{0.109}  \\
    \hline
    \multicolumn{9}{c}{\textit{Methods based on View Synthesis}} \\  % 合并两列并居中
    \hline
    COGS~\cite{jiang2024cogs}  & 0.411  & 0.172  & 0.378  & 0.131  & 0.398  & 0.121  & 0.396  & 0.142  \\
    SD-Inpaint~\cite{yu2023SDinpainting} & 0.269  & 0.164  & 0.271  & 0.142  & 0.241  & 0.114  & 0.260  & 0.140  \\
    GenWarp~\cite{seo2024genwarp} & 0.264  & 0.152  & 0.251  & 0.130  & 0.234  & 0.108  & 0.250  & 0.130  \\
    \textbf{DMAligner} (Ours)   & \underline{0.246}  & \textbf{0.123} & 0.186  & 0.117  & \textbf{0.202} & \textbf{0.084} & \textbf{0.211} & \textbf{0.108} \\
    \bottomrule
    \end{tabular}
  \label{tab:compare_sintel_davis}
  \vspace{-1.8em}
\end{table*}

\textbf{Loss Function:} Our model directly learns to denoise and reconstruct a latent representation $\tilde{x}_0$ from $x_t$ that matches the distribution of $V_{gt}$. Then, it is decoded by $\mathcal{D}$ into the RGB image $\tilde{I}_{pred}$, supervised by $I_{gt}$ in the image domain to ensure reliable and accurate output. Additionally, the trainer learns the motion foreground mask $\tilde{M}_{pred}$, which is then upsampled to obtain $\hat{M}_{pred}$, helping the model to focus on regions with significant variance. The denoising loss function, namely Denoising Loss, is defined as follows:
\begin{equation}
        \hat{M}_{pred} = Upsample\{d_r(\tilde{M}_{pred})\},
\end{equation}
\begin{equation}
\begin{aligned}
    L_{\rm{Denoising}}&=(1-\gamma )\left \|(1-\hat{M}_{pred})(I_{gt}-I_{pred}) \right \|_2 \\
                    & + \gamma  \left \| (\hat{M}_{pred})(I_{gt}-I_{pred}) \right \|_2 ,
\end{aligned}
\label{mseloss}
\end{equation}
where $\gamma$ is a variable weighting coefficient, and $d_r$ represents the dilation operation applied to the mask $r$ times. We also supervise the mask $\tilde{M}_{pred}$ using the cross-entropy loss, referred to as the Mask Loss:
\begin{equation}
\begin{aligned}
    L_{\rm{Mask}}=-M_{gt}\log(\hat{M}_{pred})-(1-M_{gt})\log(1-\hat{M}_{pred}).
\end{aligned}
\label{maskloss}
\end{equation}
The total loss is:
\begin{equation}
L_{\rm{Total}}=\lambda_1 L_{\rm{Denoising}} + \lambda_2 L_{\rm{Mask}}.
\label{totalloss}
\end{equation}
% In experiments, we set the parameters $r$ of $d_r$ is 10, $\lambda_1$ is 2 and $\lambda_2$ is 0.1.

In our experiments, the downsampling factor of $s$, the parameters $r$ (in $d_r$), $\lambda_1$, and $\lambda_2$ are set to 4, 2, 2, and 0.1, respectively.

\subsubsection{Dynamics-aware Denoised Inference.}
After training, the diffusion model $\theta$ performs inference by denoising a standard Gaussian noise $x_{T} \in \mathcal{N}(0,1)$, conditioned on $V_1$, $V_2$ and $V_M$. This process reconstructs a reliable and accurate alignment result through recurrent denoising. Specifically, the model $\theta$ takes $x_{T}$, the condition, and the timestep $T$ as inputs according to Eq.~(\ref{forwardpredx0}), producing an initial coarse denoised output $\hat{x}_0^{(T)}$. Then, following the non-Markovian forward process~\cite{song2021ddim} parametrized by $\sigma$ as shown in Eq.~(\ref{revsersext}, it generates $x_{T-1}$ at timestep $T-1$. This process is repeated recurrently, producing a sequence of progressively refined denoised results $\hat{x}_0^{(T)}, \hat{x}_0^{(T-1)}, \cdots, \hat{x}_0^{(1)}$. Finally, $\hat{x}_0^{(1)}$ is passed through the decoder $\mathcal{D}$ to output the aligned image $I_{pred}$. The non-Markovian forward process is defined as:
\begin{equation}
x_{t-1}=\sqrt{\alpha_{t-1}}\hat{x}_0^{(t)}+\sqrt{1-\alpha_{t-1}-\sigma_{t}^2} \hat{\epsilon}_{t} + \sigma_{t}x_{T},
\label{revsersext}
\end{equation}
where $\hat{x}_0^{(t)}$ is the prediction of $x_0$ from $x_t$ using the diffusion model $\theta$, and $\hat{\epsilon}_{t}$ is the approximate noise calculated by
% Eq.~\ref{revsersenoise}.
\begin{equation}
\hat{\epsilon}_{t}=\frac{x_t-\sqrt{\alpha_t}\hat{x}_0^{(t)}}{\sqrt{1-\alpha_t}}.
\label{revsersenoise}
\end{equation}
In the recurrent update process described above, the step size $\Delta$ is not fixed and can be adjusted to other values. By setting an appropriate step size $\Delta$, inference can be accelerated while still allowing incremental refinement of the results. The iterative update process can be expressed as: $\{x_T \to \hat{x}_0^{(T)}\} \to \{x_{T-\Delta} \to \hat{x}_0^{(T-\Delta)}\} \to \cdots \to \{x_1 \to \hat{x}_0^{(1)}\}$. In our empirical analysis, we set $T=1000$ and $\Delta=20$.

\section{Experiments}
\subsection{Input Structure}

As detailed in Tab.~\ref{tab:input_comparison}, our DMAligner and DPFlow~\cite{Morimitsu2025DPFlow} are highly efficient, requiring only two consecutive frames ($I_1, I_2$) for alignment-oriented view synthesis and optical flow estimation, respectively.
In contrast, 3DGS-based methods heavily depend on auxiliary priors for scene reconstruction. Beyond two input images ($I_1$ and $I_2$), COGS~\cite{jiang2024cogs} requires depth maps ($D_1, D_2$) estimated by Marigold~\cite{ke2024marigold} and foreground masks ($M_1, M_2$) predicted by FC-CLIP~\cite{yu2023fcclip}. AccidentalGS~\cite{mao2025accidentalgs} similarly relies on two input images ($I_1$ and $I_2$), depth priors ($D_1, D_2$) and camera intrinsics ($K$).
Furthermore, while GenWarp~\cite{seo2024genwarp} claims single-image input ($I_2$), it implicitly requires both $I_1$ and $I_2$ to estimate relative camera motion $P_{I_2 \to I_1}$ via optical flow~\cite{Luo_2026_EEMFlow,Luo_2024_EEMFlow,liu2016meshflow}, alongside camera intrinsics ($K$) and depth ($D_2$) from Marigold~\cite{ke2024marigold}. Ultimately, unlike these geometry-dependent approaches, our DMAligner achieves accurate alignment without requiring depth, masks, intrinsics, or pose transformations, offering a much simpler and more direct framework.

\subsection{Implementation Details}
% \subsubsection{Datasets.} 
Our DSIA dataset comprises 1,033 indoor and outdoor scenes with a resolution of 960$\times$540. The dataset is divided into a training set with over 30k samples and a test set with 2k samples. %To comprehensively evaluate the performance of various image alignment algorithms and to investigate the impact of different motion magnitudes on alignment results, we partition the test set into four subsets based on the magnitude of camera and foreground motion: LcLf (large camera motion, large foreground motion), LcSf (large camera motion, small foreground motion), ScLf (small camera motion, large foreground motion), and ScSf (small camera motion, small foreground motion). 
We partition the test set into four subsets based on the magnitude of camera and foreground motion: LcLf, LcSf, ScLf and ScSf (Lc: large camera motion, Lf: large foreground motion, Sc: small camera motion, Sf: small foreground motion). 
For the training phase, we initialize a VQ-VAE with pre-trained weights from LDM~\cite{rombach2022ldm} and fine-tune it on our DSIA dataset. Subsequently, we freeze the VQ-VAE encoder and \textit{train our custom denoising UNet from scratch} to address the alignment task. Training uses random $480\times256$ crops with flipping, while inference supports larger inputs on a single GPU.

% We conduct experiments on the DSIA dataset, the synthetic Sintel dataset~\cite{butler2012sintel}, and the real-world DAVIS dataset~\cite{caelles2019DAVIS}.
% For DSIA, with ground truth labels, we evaluate PSNR and SSIM between the predicted $I_{pred}$ and the ground truth $I_{gt}$.
% For Sintel and DAVIS, which lack ground truth, we compute LPIPS~\cite{zhang2018lpips} and DreamSim~\cite{fu2023dreamsim} between $I_{pred}$ and the reference image $I_1$ following Realfill~\cite{Tang2024realfill}.

% \subsubsection{Training Details.} For training DMAligner, we utilize a batch size of 6 with the NVIDIA A100 GPU, employing the AdamW optimizer with a one-cycle learning rate using PyTorch. Following LDM~\cite{rombach2022ldm}, we use VQ-regularized first-stage model and load publicly available pretrained weights to convert the image domain into the latent representational space. After fine-tuning the first-stage model, we keep it frozen and only train the diffusion model with the latent representations. We perform random cropping to 480$\times$256 and apply vertical or horizontal flipping during training. In inference, a single GPU can handle inputs larger than 480$\times$256.

\subsection{Comparison with Existing Methods}

% \subsubsection{Results on DSIA.} In Tab.~\ref{tab:compair}, we compare the performance of DMAligner trained on the DSIA dataset, tested on four subsets (LcLf, LcSf, ScLf, ScSf) using PSNR and SSIM metrics. We also provide an ``Avg" column for the overall performance. We fine-tuned several top optical flow networks on the DSIA dataset for comparison, using their estimated optical flow for backward warping to align images. The results show that DMAligner outperforms conventional alignment methods. It achieves a 13.7\% improvement in the average PSNR, increasing from 21.83dB to 24.82dB, and maintains leading SSIM performance. 
% DMAligner also excels in challenging scenarios with large motion. For example, in the ``LcLf" subset, DMAligner improves PSNR by 19.4\%, from 20.83dB to 24.88dB, and SSIM by 6.0\%, from 0.83 to 0.88. Even in small displacement scenarios, it shows significant gains, with an 11.4\% PSNR increase in the ``ScLf" subset.

\subsubsection{Results on DSIA.} Tab.~\ref{tab:compair} compares our DMAligner with some flow-warping methods and view-synthesis methods. In our DSIA dataset, all flow-warping methods are trained using rendered ground truth optical flow. We then align $I_2$ to $I_{gt}$ using back warping and compute PSNR and SSIM between $I_{pred}$ and $I_{gt}$. To address occlusion and ghosting artifacts from warping, these regions (determined by forward-backward consistency check~\cite{xu2022gmflow}) are excluded from the metrics calculation, as marked by $^{\dagger}$. Fig.~\ref{fig:comparison} shows qualitative results on the DSIA test set, where our DMAligner most closely matches the ground truth.

\subsubsection{Results on Sintel and DAVIS.} 
% We conducted qualitative comparisons on the real-world DAVIS~\cite{caelles2019DAVIS} dataset and the synthetic Sintel~\cite{butler2012sintel} dataset to assess the cross-domain performance of DMAligner and DSIA dataset. 在DSIA数据集上训练好所有对比方法和我们的DMAligner，我们直接在Sintel Clean和Final训练集和DAVIS数据集上测试对齐结果，使用LPIPS和DreamSim计算$I_{pred}$和$I_1$的feature distance。对于COGS所需的mask和depth，我们使用FC-ClIP和Marigold估计得到。对于SD-Inpainting和GenWarp，我们使用FlowDiffuser估计的光流和Marigold估计的depth作为输入。表二展示了在Sintel和DSIA的定量对比结果，我们的DMAligner在平均的LPSPS和DreamSim都达到了最好的表现。Fig.~\ref{fig:comparison_davis} presents examples from the DAVIS dataset. Fig.~\ref{fig:comparison_sintel} shows results from the Sintel test set. The results indicate that DMAligner demonstrates outstanding image alignment capabilities in both virtual and real-world scenarios. 
We conduct qualitative comparisons on the synthetic Sintel~\cite{butler2012sintel} dataset and the real-world DAVIS~\cite{caelles2019DAVIS} dataset to evaluate the cross-domain performance of DMAligner and DSIA dataset. All competing methods and our DMAligner are trained on DSIA, and alignment results are directly tested on the Sintel Clean, Sintel Final training sets, and the DAVIS dataset. As there is no ground truth for precise evaluation, we calculate the feature distances between $I_{pred}$ and $I_1$ using LPIPS and DreamSim~\cite{fu2023dreamsim}. Tab.~\ref{tab:compare_sintel_davis} presents the quantitative results on Sintel and DAVIS, where DMAligner achieves the best average LPIPS and DreamSim scores. 
Fig.~\ref{fig:comparison_davis} shows examples from the DAVIS, and Fig.~\ref{fig:comparison_sintel} displays results from the Sintel. These results demonstrate that DMAligner delivers exceptional image alignment performance in both synthetic and real-world scenarios.

%Since DAVIS lacks optical flow labels, we used the publicly available weights for RAFT and FlowDiffuser. The results show that RAFT and FlowDiffuser generate significant ghosting effects, leading to blurry and distorted alignments, such as in the second example with the girl's arm and the third example with the man's arm. DMAligner, however, avoids these issues and delivers clean alignment results. Fig.~\ref{fig:comparison_sintel} shows results from the Sintel test set. RAFT, FlowDiffuser, and Meshflowstruggle with black borders, holes from occlusion, and ghosting due to overlapping foregrounds. For instance, the second example shows noticeable ghosting that Meshflowfails to fix. 
%These experiments validate DMAligner's strong image alignment abilities and highlight the cross-domain versatility of the DSIA dataset, showing their effectiveness across a wide range of real-world scenarios.
\subsubsection{Results on Downstream HDR Task.} 
In HDR imaging~\cite{xu2024hdrflow}, LDR frames with alternating exposures (e.g., $L_{t-1}$,$L_{t}$,$L_{t+1}$) are aligned using flow warping before fusion. We replace this step with DMAligner, which aligns foreground and background more robustly, generating warped frames ($L_{t+1 \to t}$, $L_{t-1 \to t}$). Importantly, the fusion network and overall pipeline remain unchanged—only the aligned inputs differ. In Fig.~\ref{fig:hdr_compairison}, our method produces fewer artifacts, demonstrating strong generalization with existing HDR fusion networks.

% \subsection{Ablation Study} In Tab.~\ref{tab:ablation}, we explore the impact of predict methods and DMP. Comparing (a)\&(b) and (c)\&(d)\&(e), ``predict $x_0$" is more effective for image alignment, as it directly targets image reconstruction. comparing (b)\&(e), DMP resultes in more accurate alignment, as it allows the model to better focus on relevant areas.

\begin{table}[t]
  \caption{Ablation studies of predict targets and our DMP.}
    \vspace{-.3cm}
  \centering
    \begin{tabular}{l|c|c|cc}
    \toprule
    \multirow{2}{*}{Exp.} & Predict & \multirow{2}{*}{DMP}  & \multicolumn{2}{c}{Avg} \\
          & Targets          &  & PSNR  & SSIM \\
    \hline
    (a)   &pred $\hat{\epsilon}_t$     &w/o      & 24.82 & 0.77 \\
    (b)   &pred  $x_0$   &w/o        & 25.12 & 0.79 \\
    (c)   &rectified flow   &w/    & 25.98 & 0.80 \\
    (d)   &v-prediction     &w/      & 26.16 & 0.81 \\
    (e)  &pred $x_0$    &w/        & \textbf{26.67} & \textbf{0.81} \\
    \bottomrule
    \end{tabular}
  \label{tab:ablation}
  \vspace{-.6cm}
\end{table}

% \vspace{.5em}
% \noindent \textbf{Ablation Study.} In Tab.~\ref{tab:ablation}, we explore the impact of predict methods and DMP. Comparing (a)\&(b) and (c)\&(d)\&(e), ``predict $x_0$" is more effective for image alignment, as it directly targets image reconstruction. comparing (b)\&(e), DMP resultes in more accurate alignment, as it allows the model to better focus on relevant areas. 
\noindent \textbf{Ablation Study.} Tab.~\ref{tab:ablation} ablates prediction targets and DMP. Experiments (a)-(e) show that ``predict $x_0$" achieves better alignment by directly targeting image reconstruction. Besides, (b) vs. (e) demonstrates that DMP enhances alignment by focusing the model on relevant dynamic regions.

\begin{figure}
    \centering
    \includegraphics[width=0.9\linewidth]{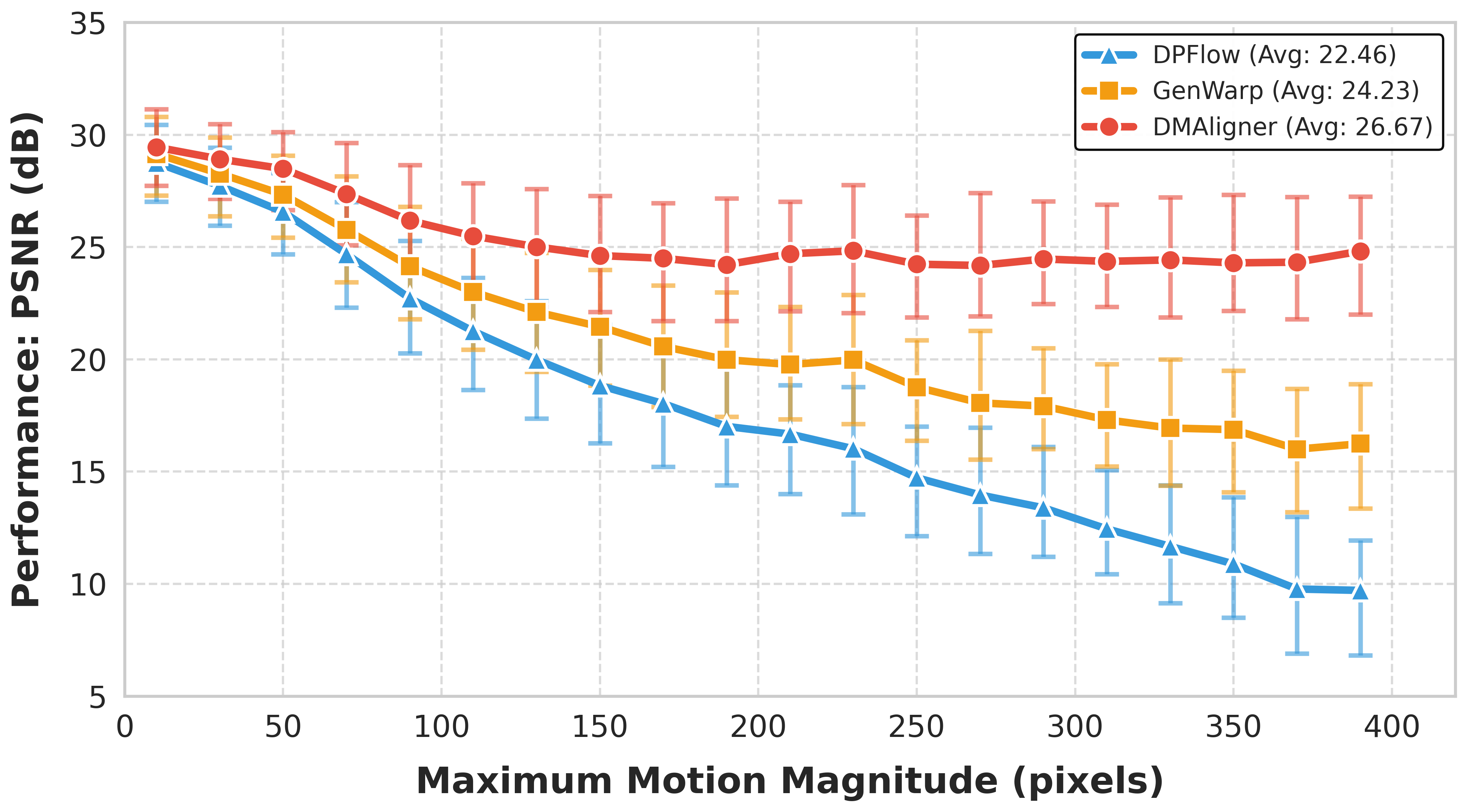}
    \vspace{-.4cm}
    \caption{Stress test: performance vs. motion magnitude.}
    \label{fig:stress_test}
    \vspace{-.9cm}
\end{figure}

% \vspace{.5em}
\noindent \textbf{Robustness Analysis.} Stress tests in Fig.~\ref{fig:stress_test} reveal that DMAligner maintains a slower degradation rate and outperforms others even under large motion magnitude gaps.

\vspace{-.1cm}
\section{Conclusion}
\vspace{-.1cm}
This work represents a significant paradigm shift in image alignment by introducing a generation-based solution. The proposed DMAligner framework, a diffusion-based approach for image alignment through alignment-oriented view synthesis, effectively addresses challenges such as ghosting artifacts, occlusions, and illumination changes that previous methods struggle with. Notably, a Dynamics-aware Mask Producing (DMP) module is introduced to enable the diffusion model to better handle dynamic foreground regions that are prone to errors. Additionally, the Dynamic Scene Image Alignment (DSIA) dataset is developed specifically for training. Extensive experiments demonstrate that DMAligner outperforms previous methods with strong cross-domain capabilities, excelling in both synthetic and real-world scenarios.

% \vspace{.5em}
\noindent \textbf{Acknowledgements.} This work was supported by Sichuan Science and Technology Program under grant No.~2025ZYD0181, in part by the National Natural Science Foundation of China (NSFC) under grant Nos.~62372091 and 62402402, in part by the Hainan Province Science and Technology Plan Project under Grant ZDYF2024(LALH)001, and in part by the Fundamental Research Funds for the Central Universities under grant No. 2682025CX116.

{
    \small
    \bibliographystyle{ieeenat_fullname}
    \bibliography{main}
}

% WARNING: do not forget to delete the supplementary pages from your submission 
% \input{sec/X_suppl}

\end{document}